\documentclass[letterpaper]{article} 
\usepackage{aaai2026}  
\usepackage{times}  
\usepackage{helvet}  
\usepackage{courier}  
\usepackage[hyphens]{url}  
\usepackage{graphicx} 
\usepackage{natbib}  
\usepackage{caption} 
\usepackage{algorithm}
\usepackage{algorithmic}

\usepackage{newfloat}
\usepackage{listings}
\DeclareCaptionStyle{ruled}{labelfont=normalfont,labelsep=colon,strut=off} 
\floatstyle{ruled}
\newfloat{listing}{tb}{lst}{}
\floatname{listing}{Listing}
\title{iCD: A Implicit Clustering Distillation Mathod for Structural Information Mining}
\author{
Xiang Xue\textsuperscript{\rm 1},
Yatu Ji\textsuperscript{\rm 1,}\thanks{Corresponding author.},
Qing-dao-er-ji Ren\textsuperscript{\rm 1},
Bao Shi\textsuperscript{\rm 1},
Min Lu\textsuperscript{\rm 1},
Nier Wu\textsuperscript{\rm 1},
Xufei Zhuang\textsuperscript{\rm 1},\\
Haiteng Xu\textsuperscript{\rm 1},
Gan-qi-qi-ge Cha\textsuperscript{\rm 2}
}
\affiliations{
\textsuperscript{\rm 1}Inner Mongolia University of Technology, China\\
\textsuperscript{\rm 2}Inner Mongolia Autonomous Region Water Conservancy Development Center, China\\
\{20231100134, MLjyt, renqingln,kshibao, cslumin, wunier04, zxf, 20241100135\}@imut.edu.cn, 1733860628@qq.com

}

\usepackage{bibentry}

\begin{document}

\maketitle

\begin{abstract}
Logit Knowledge Distillation has gained substantial research interest in recent years due to its simplicity and lack of requirement for intermediate feature alignment; however, it suffers from limited interpretability in its decision-making process. To address this, we propose \textbf{i}mplicit \textbf{C}lustering \textbf{D}istillation (iCD): a simple and effective method that mines and transfers interpretable structural knowledge from logits, without requiring ground-truth labels or feature-space alignment. iCD leverages Gram matrices over decoupled local logit representations to enable student models to learn latent semantic structural patterns. Extensive experiments on benchmark datasets demonstrate the effectiveness of iCD across diverse teacher-student architectures, with particularly strong performance in fine-grained classification tasks---achieving a peak improvement of +5.08\% over the baseline. The code is available at: https://github.com/maomaochongaa/iCD.
\end{abstract}
\section{Introduction}
Knowledge distillation, which enhances the generalization capability of student networks by transferring implicit knowledge embedded in pretrained teacher models, has become a pivotal technique in model compression \cite{gou2021knowledge,gao2025fairness}. Depending on the knowledge transfer locations, these methods are generally categorized into two groups: logit-based distillation \cite{hinton2015distilling} and feature-based distillation \cite{romero2015fitnets}. Among them, logit-level knowledge transfer has gained substantial attention in heterogeneous model compression tasks in recent years, owing to its low computational cost \cite{zhao2022decoupled} and cross-architecture learning capability without requiring intermediate structural alignment \cite{wei2024privileged}.


The logit distillation methods proposed to date can be broadly categorized into two groups. The first group extracts richer logit knowledge by introducing multiple classifiers \cite{yao2020knowledge,lan2018knowledge} or self-supervised learning \cite{xu2020knowledge}. The second group optimizes the knowledge transfer process through techniques like dynamic temperature \cite{xu2020feature,zheng2024knowledge} or knowledge decoupling \cite{ijcai2021p444,yang2023from,zhao2022decoupled}. While these methods achieve promising results, their sole reliance on global logit knowledge from the entire input may lead to suboptimal outcomes. To address this limitation, SDD \cite{wei2024scale} refines semantic knowledge by decoupling logit outputs at the scale level, enhancing student models’ discriminative capacity for ambiguous samples. However, logit-based distillation methods fundamentally remain "learn-to-predict" approaches---they operate in the output space rather than modeling structural relationships in the representation space.

To this end, we propose the \textbf{i}mplicit \textbf{C}lustering \textbf{D}istillation (iCD) method, which captures the teacher's structural semantic dependencies by leveraging scale-decoupled logit outputs. Specifically, iCD first decouples the entire logit output into localized region-wise logits, facilitating the acquisition of richer and more explicit semantic knowledge. Subsequently, Gram matrices are computed for both local and global logits to capture spatial structural information across granularities. These Gram matrices from spatially corresponding logit positions are then paired. By contrasting the feature distributions of teacher and student at identical positions, iCD guides the student to learn the local semantic structural organization patterns captured by the teacher, rather than merely imitating specific feature values.

Our motivation stems from the global prediction ambiguity in logit distillation, and the proposed iCD method redefines knowledge transfer by modeling structural semantic relationships among logits. Operating on the projected feature vectors from penultimate-layer feature maps, iCD constructs semantically coherent feature clusters through implicit clustering (where spatially corresponding teacher-student features are viewed as feature clusters), enabling cluster-level knowledge transfer. Unlike conventional feature imitation approaches, iCD requires no explicit spatial/channel alignment, demonstrating superior flexibility and interpretability. Extensive experiments validate the effectiveness of our method. The key contributions of this work are summarized as follows:

(1) We propose iCD, a novel approach that integrates clustering mechanisms into logit learning, enforcing structural distribution alignment for high-level feature representations between student and teacher models.

(2) iCD operates as a semi-supervised learning framework, requiring no explicit label supervision. It leverages Gram matrix representations of global and localized high-level logits from both teacher and student for clustering. This inherently encourages the student to approximate the teacher’s logit distributions during feature representation learning.

(3) Extensive experimental results on the CIFAR-100 and CUB-200-2011 datasets demonstrate that our approaches surpasses existing state-of-the-art logit-based knowledge distillation methods. The significant performance gains on fine-grained tasks empirically validate the effectiveness and rationality of clustering mechanisms in knowledge distillation.

\section{Related Work}
\subsection{Feature-based Distillation}
Intermediate feature-based knowledge distillation aims to extract rich semantic information from the deep-layer representations of teacher models to guide student model learning  \cite{chen2021distilling,guo2023class,huang2023knowledge,komodakis2017paying,liu2023function,romero2015fitnets,tian2020contrastive,yang2024vittkd}. FitNets \cite{romero2015fitnets} first proposed having students directly mimic intermediate feature map outputs from teachers, establishing the paradigm of explicit feature imitation. Subsequently, numerous studies have focused on enhancing the effectiveness and generalization capacity of feature alignment.

Methods employing attention mechanisms enhance knowledge transfer for critical regions: For instance, AT \cite{komodakis2017paying} utilizes activation intensities of feature maps to generate attention maps, guiding students to focus on semantically salient regions. Another category targets structural relationship transfer: RKD \cite{park2019relational} and SP \cite{tung2019similarity} model inter-sample geometric relationships and similarity patterns respectively, while ReviewKD \cite{chen2021distilling} enhances representation consistency by revisiting shallow-layer knowledge. CRD \cite{tian2020contrastive} introduces a contrastive learning framework that aligns teacher-student feature distributions using positive-negative sample pair objectives. FCFD \cite{liu2023function} optimizes functional similarity between features to improve imitation efficiency; VID \cite{ahn2019variational} achieves tighter semantic alignment by maximizing mutual information between teacher-student features. SemCKD \cite{wang2023semckd} automatically assigns semantically calibrated teacher layers to student layers via attention mechanisms.

Further advances have moved beyond direct feature alignment to capture richer semantic structures within the feature space. FSP matrix method \cite{yim2017gift} characterizes inter-layer feature transformation relationships, enabling knowledge transfer via inter-layer similarity matrix matching. NORM \cite{liu2023norm} introduces an N-to-one representation matching mechanism to enhance transfer efficiency without the need for explicit alignment. CoSS \cite{singh2024simple} explores spatial similarity-based unsupervised distillation to reduce label dependency. DisWOT \cite{dong2023diswot} applies knowledge distillation to zero-shot architecture search, achieving training-free student model selection.

\subsection{Logit-based Distillation}
Hinton et al. \cite{hinton2015distilling} pioneered the soft logit-based distillation framework, enabling students to mimic the probability distributions of teacher outputs. Subsequent extensive studies  \cite{li2022asymmetric,zhou2021rethinking,yang2023from,sun2024logit,xu2024improving,bao2024post} have focused on enhancing the efficiency and effectiveness of logit distillation.

From a mechanistic understanding perspective, Tang et al. \cite{tang2020understanding} demonstrated that Label Smoothing (LS) enhances student performance. Müller et al. \cite{muller2019when} demonstrated that LS reduces intra-class variation, leading to more compact feature representations, while Shen et al. \cite{shen2021label} further proved that tight clustering strengthens cross-class semantic separability. However, existing methods such as Deep Mutual Learning \cite{zhang2018deep} and Teacher Assistant  \cite{mirzadeh2020improved,son2021densely,guo2024pixel} still exhibit limited effectiveness in practical applications.

Concurrently, recent advances focus on refining logit knowledge, evolving along two key directions: 
\textbf{Dynamic Temperature Strategies.} Conventional KD employs fixed temperatures, which struggle to adapt to varying samples or task complexities. CTKD\cite{li2023curriculum} introduces learnable temperatures to modulate task difficulty; LSKD \cite{sun2024logit} dynamically adjusts temperatures based on the statistical variance of logits; MLKD \cite{jin2023multi} and WTTM \cite{zheng2024knowledge} adopt multi-temperature strategies or transformed matching to optimize knowledge transfer.
\textbf{Logit Knowledge Decoupling.} Global logits blend target/non-target class information, causing semantic ambiguity. DKD \cite{zhao2022decoupled} decouples KL divergence into target/non-target knowledge for separate optimization; NKD \cite{yang2023from} normalizes non-target logits to enhance soft label utilization; ATS \cite{li2022asymmetric} decomposes knowledge into calibration, smoothing, and discriminative components; ReKD \cite{xu2024improving} implements differentiated distillation for head/tail categories; SDD \cite{wei2024scale} decouples global/local logit outputs to boost student absorption of fine-grained knowledge.

Additional innovations include: FN \cite{xu2020feature} achieving sample-adaptive calibration via the L2 norm of features; SSKD \cite{xu2020knowledge} extracting richer latent knowledge through self-supervised learning; KDExplainer \cite{xue2021kdexplainer} introducing virtual attention mechanisms to mitigate class conflicts; TeKAP \cite{hossain2025single} simulating multi-teacher distillation by perturbing feature maps and logits of a single pre-trained teacher to generate diverse teacher perspectives; LDRLD \cite{xu2025local} recursively decoupling and reorganizing logit information to capture inter-class relationships.

While the aforementioned methods achieve compelling results, they largely overlook the structural information inherent in the logits of the teacher and student models. To address this, we propose iCD, which employs implicit clustering to enforce structural alignment between their logit distributions through cluster-level knowledge transfer.

\section{Methodology}
In this section, we first revisit traditional Knowledge Distillation (KD) and Scale-Decoupled Knowledge Distillation (SDD), and then present our proposed implicit Clustering Distillation (iCD).

\textbf{Notation.}
Given an input image \( x \), let \( T \) and \( S \) denote the teacher network and student network, respectively. We decompose each network into two components:
(i) A convolutional feature extractor \( f_{Net} \), where \( {Net} = \{T, S\} \). The feature map from the penultimate layer is represented as \({f}_{Net}(x) \in {R}^{{c}_{Net} \times {h}_{Net} \times {w}_{Net}} \), where \( c_{Net} \) is the number of feature channels, and \( h_{Net} \times w_{Net} \) denotes the spatial dimensions.
(ii) A projection matrix \( W_{{Net}}(x) \in {R}^{{c}_{Net} \times K} \) that maps the feature vectors extracted from \( {f}_{Net}(x) \) to logits \( z^l_{{Net}} \) (\( l = 1,2,\ldots,K \)) corresponding to \( K \) categories.
Let  \({f}_{Net}(i,j) = {f}_{Net}(x)(:j,k) \in {R}^{c_{Net} \times 1 \times 1} \) denote the feature vector at spatial position \( (j,k) \) in \( {f}_{Net}(x) \). According to the receptive field theory in \cite{he2015spatial}, \( f_{Net}(j,k) \) corresponds to the representation of the region \( (t_x, t_y, t_x + d, t_y + d) \) in \( x \), where \( t_x = d \cdot j \), \( t_y = d \cdot k \), and \( d \) is the downsampling factor between the input and the final feature map. Define all scales $m$ in the set $M = \{1, 2, 4, \ldots, w\}$, where each scale $m$ partitions the feature map into $N_m = m^2$ non-overlapping cells.

\begin{figure*}[htbp]
\centering
\includegraphics[width=0.8\textwidth]{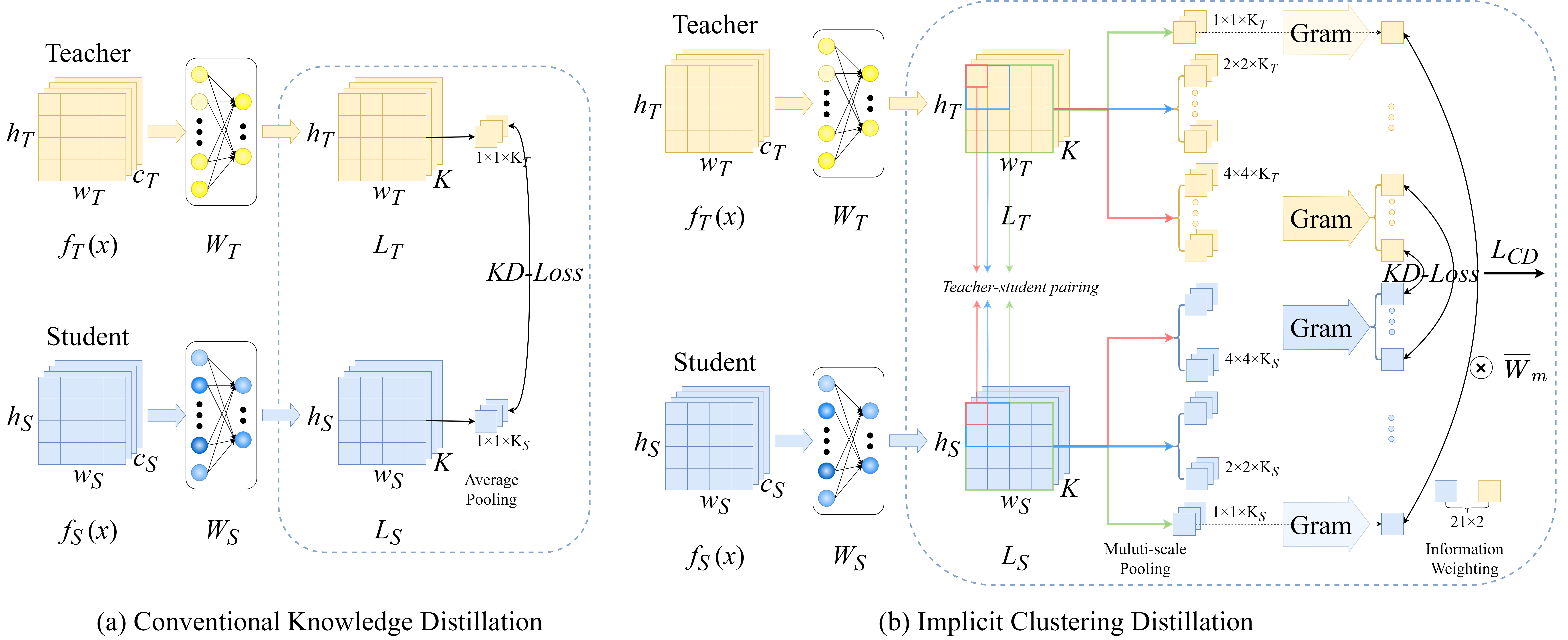} 
\caption{Visualization of Knowledge Distillation and Implicit Clustering Distillation.}
\label{fig:att}
\end{figure*}

\subsection{KD and SDD}
KD, initially introduced in \cite{hinton2015distilling}, aims to transfer and distill knowledge embedded in the teacher model's logits to the student model through specialized loss functions.

\begin{eqnarray}
L_{\mathcal{LD}} &=& \mathcal{KL}(\sigma(P_T) \| \sigma(P_S)) \label{eq:1} \\
P_T &=& W_T \sum_{j=0}^{h_T-1} \sum_{k=0}^{w_T-1} \frac{1}{h_T w_T} f_T(j,k) \nonumber \\
    &=& \sum_{j=0}^{h_T-1} \sum_{k=0}^{w_T-1} \frac{1}{h_T w_T} L_T(j,k) \label{eq:2} \\
P_S &=& W_S \sum_{j=0}^{h_S-1} \sum_{k=0}^{w_S-1} \frac{1}{h_S w_S} f_S(j,k) \nonumber \\
    &=& \sum_{j=0}^{h_S-1} \sum_{k=0}^{w_S-1} \frac{1}{h_S w_S} L_S(j,k) \label{eq:3}
\end{eqnarray}

Where \( \sigma(\cdot) \) denotes the softmax activation function and \( \mathcal{KL}(\cdot,\cdot) \) represents the Kullback-Leibler divergence. Given the linear property of fully-connected (FC) layers, \( P_T \) and \( P_S \) can be further simplified.
Logit outputs are defined as:\( L_S = W_S f_S(x) \) (student) and \( L_T = W_T f_T(x) \) (teacher), where \( W \) denotes FC layer weights.

This concludes the description of traditional logit-based distillation methods. We now briefly introduce their enhanced counterpart: SDD.

SDD enhances the traditional KD framework by replacing its single global average pooling with multi-scale pooling operations, thereby preserving fine-grained knowledge with well-defined semantics to facilitate student learning. It further incorporates feature weighting to direct the student network's focus toward ambiguous samples where local and global class predictions diverge.

Let \( {C}(m,n) \) denote the cell grid for the \( n \)-th cell at the \( m \)-th scale, \( \mathcal{Z}(m,n) \) represent the corresponding input region, and \( \pi(m,n) \in {R}^{K \times 1 \times 1} \) indicate the logit output. For region \( \mathcal{Z}(m,n) \), the aggregated logits from teacher and student models is formulated in Eq. \ref{eq:4} and \ref{eq:5}.
\begin{eqnarray}
    \pi_T(m,n) = \sum_{j,k \in \mathcal{C}(m,n)} \frac{1}{m^2} L_T(j,k) \label{eq:4} \\
    \pi_S(m,n) = \sum_{j,k \in \mathcal{C}(m,n)} \frac{1}{m^2} L_S(j,k) \label{eq:5}
\end{eqnarray}

Here, \(( j,k) \) denote coordinates of logit outputs in \( \mathcal{C}(m,n) \). For each logit output pair, the distillation loss \( \mathcal{D}(m,n) \) encoding logit knowledge transfer from teacher to student across region \( \mathcal{Z}(m,n) \) is defined by Eq. \ref{eq:6}.
\begin{eqnarray}
    \mathcal{D}(m,n) &=& \mathcal{LD}\left(\sigma(\pi_T(m,n)), \sigma(\pi_S(m,n))\right) \label{eq:6} \\
    \mathcal{L}_{SDD} &=& \sum_{m \in M} \sum_{n \in N} \mathcal{D}(m,n)  \label{eq:7}
\end{eqnarray}

Here, \( \mathcal{LD}(\cdot,\cdot) \) represents traditional logit-based distillation losses including the Kullback-Leibler divergence in \cite{hinton2015distilling} and decoupled loss in \cite{zhao2022decoupled}. By iterating through all scales \( m \) in set \( M=\{1,2,4,\ldots,w\} \) with their corresponding total cells \( N_m=\{1,4,16,\ldots,w^2\} \), the final SDD loss is derived as formulated in Eq. \ref{eq:7}.

Traditional logit-based distillation methods leverage only averaged logit outputs that blend diverse local logit knowledge as shown in the above equations. Distinct local outputs in this paradigm inherently contain unique semantic information. Their naive fusion into a single logit output introduces ambiguous knowledge transfer to the student model and misguides its learning. To resolve this limitation, researchers proposed SDD which decouples logit outputs at the scale level to extract richer and more definitive logit knowledge for student model training achieving remarkable improvements. While the method generates numerous multi-scale logit patches, these patches warrant more exhaustive utilization to maximize their pedagogical value.

To fully leverage the representational capacity of multi-scale information, we propose iCD, an implicit clustered knowledge distillation method. Specifically, iCD leverages Gram matrices to mine structural representations of logits, treating teacher-student feature pairs at each scale and spatial location as clusters. This facilitates the transfer of the teacher’s structural and semantic knowledge \cite{dong2023diswot} from the teacher’s fine-grained representations to the student model.

\begin{table*}[ht]
    \centering
    \begin{tabular}{c c c c c}
        \hline
        Teacher & ResNet32x4 & WRN40-2 & ResNet50 & ResNet50 \\
        Acc     & 79.42      & 75.61   & 75.61    & 79.34    \\
        \hline
        Student & MobileNetV2 & VGG8 & MobileNetV2 & ShuffleNetV1 \\
        Acc     & 64.60      & 70.36  & 64.60      & 70.50    \\
        \hline
        FitNet  & 65.61      & 70.98  & 65.12      & 72.03    \\
        SP      & 67.52      & 73.18  & 66.34      & 71.28    \\
        CRD     & 69.13      & 73.88  & 68.89      & 75.70    \\
        SemCKD  & 68.99      & 73.67  & 68.34      & 75.56    \\
        ReviewKD& -          & -      & -          & -        \\
        MGD     & 68.13      & 73.33  & 68.55      & 74.99    \\
        \hline
        KD      & 67.72      & 73.97  & 68.87      & 75.82    \\
        SD-KD   & 68.84      & 74.44  & 69.81      & \textbf{76.87}    \\
        iCD-KD& \textbf{69.46 (+1.74)} & \textbf{74.64 (+0.77)} & \textbf{69.83 (+0.96)} & 76.54 (+0.72) \\
        DKD     & 68.98      & 74.13  & 69.33      & 77.01    \\
        SD-DKD  & 70.08      & 74.58  & 70.13      & 78.11    \\
        iCD-DKD& \textbf{70.31 (+1.33)} & \textbf{75.14 (+1.01)} & \textbf{70.42 (+1.09)} & \textbf{78.20 (+1.19)} \\
        NKD     & 68.81      & 73.80  & 68.85      & 76.22    \\
        SD-NKD  & 69.59      & 74.34  & 70.04      & \textbf{76.95}    \\
        iCD-NKD& \textbf{69.62 (+0.81)} & \textbf{74.63 (+0.83)} & \textbf{70.67 (+1.82)} & 76.65 (+0.43) \\
        \hline
    \end{tabular}
    \caption{Performance of models on the CIFAR-100 dataset. Here, the teacher and student models have different network structures and layers. Specifically, the layers of ResNet32x4, WRN40-2, and ResNet50 are 4, 4, and 3 while the layers of MobileNetV2, VGG8, and ShuffleNetV1 are 3, 3, and 4, respectively. The “iCD-” denotes the proposed iCD method integrated into baseline approaches. The “SD-” refers to the SDD method. The meanings of “iCD-” and “SD-” remain consistent in subsequent tables.}
    \label{tab:1}
\end{table*}

\subsection{iCD}
In this section, we detail the proposed iCD. As shown in Figure~\ref{fig:att}, iCD consists of two key components: \textit{structured clustering} and \textit{information weighting}. Specifically, given feature maps \(f_T{(x)}\) and \(f_S{(x)}\) extracted from the penultimate layers of the teacher and student models, respectively, iCD forms logit pairs over corresponding spatial regions $C(m,n)$  and computes the similarity of their internal fine-grained representations. This enables the student to learn fine-grained logits with structured spatial distributions. Subsequently, we establish distillation pipelines for logits at each scale $m \in M$. Finally, information weighting reassigns weights across scales, as the logits at larger \( m \) contain richer fine-grained information whose proportion increases accordingly. This guides the student network to focus precisely on local discriminative features.

Following SDD's practice~\cite{wei2024scale}, let \( {C}(m,n) \) denote the cell grid for the \( n \)-th cell at the \( m \)-th scale, \( \mathcal{Z}(m,n) \) represent the corresponding input region, and \( \pi(m,n) \in {R}^{K \times 1 \times 1} \) denote the logits.
\begin{eqnarray}
    \pi(m,n)_T' &=& \mathrm{normalize}(\pi(m,n)_T, \mathrm{dim}) \label{eq:8} \\
    \pi(m,n)_S' &=& \mathrm{normalize}(\pi(m,n)_S, \mathrm{dim}) \label{eq:9}
\end{eqnarray}

Here, \( \mathrm{normalize}(\cdot,\cdot) \) denotes the \( L^2 \)-normalization operation along the feature dimension (dim=1).

In this work, we compute Gram matrices \( G{(m,n)} \) from \( \pi(m,n)_T' \) and \( \pi(m,n)_S' \) at each scale to capture the structural knowledge embedded in logits, as formalized in Eq.~\ref{eq:10} and~\ref{eq:11}.
\begin{eqnarray}
    G_{T}(m,n) = \pi(m,n)_T'^\top \pi(m,n)_T' \label{eq:10} \\
    G_{S}(m,n) = \pi(m,n)_S'^\top \pi(m,n)_S' \label{eq:11}
\end{eqnarray}

At each scale \( m \), we compute the KL divergence between the structural distributions modeled by \( G_T{(m,n)} \) and \( G_S{(m,n)} \) for each  \( n \). 
\begin{eqnarray}
    \mathcal{D}(m,n) = \mathcal{LD}(\sigma(G_T(m,n)), \sigma(G_S(m,n))) \label{eq:12}
\end{eqnarray}

Since logit outputs at different scales contain varying levels of fine-grained knowledge, applying uniform weights is suboptimal. Therefore, we assign higher weights to finer-grained scales to guide the student model toward learning more discriminative structural patterns, as specified in Eq. \ref{eq:13}.
\begin{eqnarray}
    \overline{W}_{m} = \frac{i}{\sum_{i=1}^{|M|} i} \label{eq:13}
\end{eqnarray}

Here, \( |M| \) represents the number of elements in the set \(M\). We compute the weighted sum over all \( \mathcal{D}(m,n) \). 
\begin{eqnarray}
    \mathcal{L}_{iCD} &=& \sum_{m \in M} \sum_{n \in N} \overline{W}_m \cdot \mathcal{D}(m,n) \label{eq:14}
\end{eqnarray}

Finally, we define the total training loss in Eq. \ref{eq:15}, which combines label supervision, decoupled knowledge, and leverages the teacher model’s multi-scale structural information to enhance the student model’s performance.
\begin{eqnarray}
    \mathcal{L} &=& \mathcal{L}_{CE} + \alpha \mathcal{L}_{SDD} + \gamma \mathcal{L}_{iCD} \label{eq:15}
\end{eqnarray}

Here, \( \mathcal{L}_{CE}(\cdot) \) denotes the label supervision loss for the current task, \( \mathcal{L}_{SDD}(\cdot) \) denotes the scale-decoupling loss, and \( \alpha \) and \( \gamma \) are hyperparameters that balance their respective contributions.

\textbf{Compared with conventional Knowledge Distillation.} Traditional KD methods primarily rely on the teacher model's final logit outputs, which are amalgams of global semantic information lacking spatial or semantic decoupling. This results in ambiguous knowledge transfer to the student, hindering its capacity to comprehend intricate semantic structures. In contrast, our proposed iCD method repositions knowledge transfer from final logit predictions to logit structural exploration. It explicitly reveals and transmits the teacher model's inherent local semantic consistency and regional correlations within, thereby providing the student with clearer structural guidance. This enables learning of deeper semantic representations rather than mere imitation of final classification decisions.

\textbf{Compared with SDD.} The SDD method spatially decouples logits, decomposing global logit outputs into multiple region-specific logits to provide finer-grained supervision for student models, effectively mitigating knowledge ambiguity. However, SDD's supervisory signal remains logit-centric, with its "locality" confined to responses from specific image regions, failing to explore semantic relational structures within feature representations. In contrast, our proposed iCD method extends knowledge decoupling from logits to feature representations by structurally clustering teacher-student logit outputs. Its core strength lies in transferring semantic similarity-based relational structures within feature spaces, exposing intrinsic organizational principles of decision-making rationales rather than location-specific classification scores. This equips the student model with more interpretable and generalizable structured knowledge, enabling deeper semantic representation learning.

\begin{table*}[ht]
    \centering
    \begin{tabular}{c c c c c c c}
        \hline
        Teacher & ResNet32x4 & WRN40-2 & ResNet50 & VGG13 & ResNet32 & ResNet50 \\
        Teacher Acc & 79.42 & 75.61 & 79.34 & 74.64 & 79.42 & 79.34 \\
        \hline
        Student & ShuffleNetV1 & ShuffleNetV1 & MobileNetV2 & MobileNetV2 &ShuffleNetV2 & VGG8 \\
        Student Acc & 70.50 & 70.50 & 64.60 & 64.40 & 71.82 & 70.36 \\
        \hline
        FitNet & 73.54 & 73.73 & 63.16 & 64.14 & 73.54 & 70.69 \\
        RKD & 72.28 & 72.21 & 64.43 & 64.52 & 73.21 & 71.50 \\
        SP & 73.48 & 74.52 & 68.08 & 68.08 & 74.56 & 73.34 \\
        CRD & 75.11 & 76.05 & 69.11 & 69.11 & 75.65 & 74.30 \\
        SemCKD & 76.31 & 76.06 & 68.69 & 69.98 & 77.02 & 74.18 \\
        ReviewKD & 77.45 & 77.14 & 69.89 & 70.37 & 77.78 & 75.34 \\
        DiffKD NIPS2023 & 76.57 & - & 69.21 & - & 77.52 & - \\
        LSKD CVPR2024 & - & - & 69.02 & 68.61 & 75.56 & 74.42 \\
        WTTM ICLR2024 & 74.37 & 75.42 & 69.59 & 69.16 & 76.55 & 74.82 \\
        LDRDL ICCV2025 & 76.46 & 77.09 & 70.74 & 70.11 & 77.33 & 75.49 \\
        TeKAP ICLR2025 & 74.92 & 76.75 & 69.00 & 67.39 & 75.43 & 74.37 \\
        \hline
        KD & 74.07 & 74.83 & 67.35 & 67.37 & 74.45 & 73.81 \\
        SD-KD & 76.30 & 76.65 & 69.55 & 68.79 & 76.67 & \textbf{74.89} \\
        iCD-KD & \textbf{76.56 (+2.49)} & \textbf{76.69 (+1.86)} & \textbf{70.26 (+2.91)} & \textbf{69.43 (+2.06)} & \textbf{76.79 (+2.34)} & 74.63 (+0.82) \\
        DKD & 76.45 & 76.70 & 70.35 & 69.71 & 77.07 & 75.34 \\
        SD-DKD & 77.30 & \textbf{77.21} & 71.36 & 70.25 & \textbf{78.05} & \textbf{75.86} \\
        iCD-DKD & \textbf{77.54 (+1.09)} & 76.99 (+0.29) & \textbf{71.63 (+1.28)} & \textbf{70.36 (+0.65)} & 77.77 (+0.7) & 75.67 (+0.33) \\
        NKD & 75.31 & 75.96 & 69.39 & 68.72 & 76.26 & 74.01 \\
        SD-NKD & 76.34 & \textbf{76.81} & \textbf{70.25} & 69.50 & 77.07 & 74.62 \\
        iCD-NKD & \textbf{76.40 (+1.09)} & 76.68 (+0.72) & 69.93 (+0.54) & \textbf{69.89 (+1.17)} & \textbf{77.41 (+1.15)} & \textbf{74.69 (+0.68)} \\
        \hline
    \end{tabular}
    \caption{Performance of models on the CIFAR-100 dataset. Here, the teacher and student with different network structures but the same layer.}
    \label{tab:2}
\end{table*}

\section{Experiments}
\subsection{Experimental Setups.}
\textbf{Dataset.} Our experiments use the CIFAR-100 \cite{krizhevsky2009learning} and CUB-200 \cite{Welinder2010} datasets. CIFAR-100 is adopted for evaluating standard image classification tasks. CUB-200, which contains 200 bird species across distinct categories, is used for fine-grained image classification evaluation.

\textbf{Implementation Details.}
As described in the section on iCD, the loss function for \( \mathcal{L}_{iCD} \) can be implemented using any logit-based loss. To ensure fair comparisons with prior logit-based methods and to examine iCD's efficacy, we use the same loss functions as employed in the KD, DKD, NKD, and SSD methods, denoting these implementations as iCD-KD, iCD-DKD, and iCD-NKD respectively.

iCD contains three hyperparameters: a set of scales \( M \) and balancing parameters \( \alpha \) and \( \gamma \). We provide a detailed analysis of \( M \) and \( \gamma \) in the ablation studies. Specifically, \( M \) is set to \( \{1, 2, 4\} \) and \( \gamma=2 \). For \( \alpha \), we follow the original SDD configuration to ensure fair comparison with prior logit-based methods. Additionally, since the accumulated multi-scale logit distillation loss may induce excessively large gradient magnitudes during the early training phase, we apply a 30-epoch linear warm-up strategy in all experiments.

\textbf{Training Details.}
For CIFAR100 and CUB200 datasets, our implementation follows \cite{tian2020contrastive,wei2024scale}. Both teacher and student models are trained using SGD optimizer for 240 epochs, with batch size 64. Learning rates are set to 0.01 for ShuffleNet/MobileNet-V2 and 0.05 for other architectures (e.g., VGG, ResNet, WRN series), decaying by a factor of 10 at epochs 150, 180, and 210. Weight decay and momentum are configured as \(5\times10^{-4}\) and 0.9, respectively. Distillation loss weights align with KD, DKD, NKD, and SDD implementations to ensure comparative fairness.

\subsection{Comparison Results}
\textbf{Results on the teacher and student with different net works tructures and layers.}
As shown in Table \ref{tab:1}, the proposed iCD method demonstrates superior performance across multiple teacher-student pairs. iCD focuses on excavating and transferring local structural knowledge within individual samples from the teacher, guiding the student to learn these higher-order structural patterns. Empirical results reveal that integrating iCD into the SDD framework achieves significant and consistent improvements over baseline KD, DKD, and NKD. For instance, iCD-KD yields a +1.74\% improvement over KD in ResNet32x4→MobileNetV2 distillation, while iCD-NKD surpasses NKD by +1.82\% in WRN40-2→MobileNetV2 tasks. These gains substantiate iCD’s efficacy.

Crucially, iCD exhibits pronounced advantages in heterogeneous teacher-student distillation (e.g., ResNet→ShuffleNet/MobileNet), where architectures and layer depths differ. It not only consistently outperforms the SDD framework itself but also substantially surpasses advanced feature-based methods like FitNet \cite{romero2015fitnets}, CRD \cite{tian2020contrastive}, and SemCKD \cite{wang2023semckd}. This validates that learning semantic structural knowledge offers stronger adaptability and superiority compared to merely mimicking logit-based outputs.

\textbf{Results on the teacher and student with different net work structures but the same layer.}
As shown in Table \ref{tab:2}, iCD-KD achieves +2.49\% and +2.91\% improvements over KD in ResNet32×4→ShuffleNetV1 and ResNet50→MobileNetV2 distillation tasks, respectively. Notably, iCD-DKD significantly outperforms traditional feature distillation methods (e.g., FitNet, SP, CRD~\cite{romero2015fitnets,tung2019similarity,tian2020contrastive}) and surpasses advanced approaches including ReviewKD \cite{chen2021distilling}, MGD, and recent methods (e.g., LDRLD \cite{xu2025local}, TeKAP \cite{hossain2025single}), validating its superiority. These results confirm that iCD effectively bridges representation gaps across heterogeneous networks through fine-grained structural knowledge transfer, particularly excelling in distillation tasks involving architecturally divergent yet layer-aligned teacher-student pairs.

\begin{table*}[ht]
    \centering
    \begin{tabular}{c c c c c c}
        \hline
        Teacher & ResNet32x4 & ResNet32x4 & VGG13 & VGG13 & ResNet50 \\
        Acc     & 66.17      & 66.17      & 70.19 & 70.19 & 60.01 \\
        \hline
        Student & MobileNetV2 & ShuffleNetV1 & MobileNetV2 & VGG8 & ShuffleNetV1 \\
        Acc     & 40.23       & 37.28       & 40.23       & 46.32 & 37.28 \\
        \hline
        SP      & 48.49       & 61.83       & 44.28       & 54.78 & 55.31 \\
        CRD     & 57.45       & 62.28       & 56.45       & 66.10 & 57.45 \\
        SemCKD  & 56.89       & 63.78       & 68.23       & 66.54 & 57.20 \\
        ReviewKD& -           & 64.12       & 58.66       & 67.10 & - \\
        MGD     & -           & -           & -           & 66.89 & 57.12 \\
        LDRDL ICCV2025 & 60.99       & 65.19       & 59.73       & 68.27 & 60.46 \\
        \hline
        KD      & 56.09       & 61.68       & 53.98       & 64.18 & 57.21 \\
        SD-KD   & 60.51       & 65.46       & \textbf{59.80} & 67.32 & \textbf{60.56} \\
        iCD-KD& \textbf{61.17 (+5.08)} & \textbf{65.88 (+4.20)} & 59.61 (+5.37) & \textbf{68.52 (+4.34)} & 60.18 (+2.97) \\
        DKD     & 59.94       & 64.51       & 58.45       & 67.20 & 59.21 \\
        SD-DKD  & 62.97       & 65.58       & 64.86       & \textbf{68.67} & 60.66 \\
        iCD-DKD& \textbf{63.51 (+3.57)} & \textbf{65.95 (+1.44)} & \textbf{65.29 (+6.84)} & 68.31 (+1.11) & \textbf{60.79 (+1.57)} \\
        NKD     & 59.81       & 64.0        & 58.40       & 67.16 & 59.11 \\
        SD-NKD  & 62.69       & 65.50       & \textbf{64.63}       & 68.37 & 60.42 \\
        iCD-NKD& \textbf{63.55 (+3.71)} & \textbf{67.10 (+3.1)} & 63.70 (+5.30) & \textbf{68.95 (+1.79)} & \textbf{61.74 (+2.63)} \\
        \hline
    \end{tabular}
    \caption{Performance of models on the CUB-200 Dataset. We conducted experiments under three distinct teacher-student configurations: same architecture and layers (VGG13-VGG8), different architectures but identical layers (ResNet32×4-ShuffleNetV1 and VGG13-MobileNetV2), and both different architectures and layers (ResNet32×4-MobileNetV2 and ResNet50-ShuffleNetV1).}
    \label{tab:3}
\end{table*}
\textbf{Results on the fine-grained classification task.}
As shown in Table \ref{tab:3}, the proposed iCD method demonstrates exceptional performance across multiple teacher-student model architectures on the CUB200 fine-grained recognition dataset. iCD yields significant improvements over baseline KD, DKD, and NKD methods, especially in highly challenging heterogeneous architectures scenarios. iCD-KD achieves performance gains of +5.08\% over KD and +3.71\% over NKD in the ResNet32$\times$4$\rightarrow$MobileNetV2 transfer; iCD-NKD surpasses NKD by +2.63\% in ResNet50$\rightarrow$ShuffleNetV1; and iCD-DKD delivers a +6.84\% improvement over DKD in VGG13$\rightarrow$MobileNetV2. Notably, iCD-NKD outperforms the teacher model in the ResNet32$\times$4 $\rightarrow$ ShuffleNetV1 setting. Moreover, the iCD-KD series significantly surpasses traditional feature distillation methods (SP, CRD, SemCKD) and even outperforms the state-of-the-art LDRLD approach~\cite{xu2025local}. These results conclusively validate that iCD can effectively extract and transfer fine-grained structural knowledge, enhance student models' capacity to absorb heterogeneous knowledge, and provide robust solutions for knowledge distillation in fine-grained recognition tasks.

\subsection{Ablation Study}
To validate the effectiveness of each component in iCD, ablation studies were conducted on the CUB-200 dataset using both heterogeneous teacher-student pairs (ResNet32$\times$4 $\rightarrow$ MobileNetV2 and ResNet32$\times$4 $\rightarrow$ ShuffleNetV1) and a homogeneous pair (VGG13 $\rightarrow$ VGG8), covering diverse architectural gaps and knowledge distillation scenarios.

\textbf{Effects of Different M Settings.}
We design multiple scale combinations $M$ for iCD to investigate the impact of structural knowledge extracted from local image patches at different scales. Each scale $m \in M$ introduces structural information from region $\mathcal{Z}(m, n)$. For fair comparison, baseline methods are implemented using conventional KD. Experimental results are summarized in Table~\ref{tab:M}.
\begin{table}[ht]
    \centering
    \begin{tabular}{c c c}
        \hline
        Teacher & ResNet32x4 & VGG13 \\
        Student & ShuffleNetV1 & VGG8 \\
        \hline
        KD & 61.68 & 64.18 \\
        M=\{1\} & 62.48 & 63.31 \\
        M=\{2\} & 65.10 & 67.78 \\
        M=\{4\} & 65.48 & 67.97 \\
        M=\{1,2\} & \underline{65.74} & 67.31 \\
        M=\{1,4\} & 65.26 & 67.95 \\
        M=\{2,4\} & 65.46 & \underline{68.38} \\
        M=\{1,2,4\} & \textbf{65.88} & \textbf{68.52} \\
        \hline
    \end{tabular}
    \caption{Performance comparison of the iCD method with different multi-scale configurations. Bold indicates the best result; underlined denotes the second-best.}
    \label{tab:M}
\end{table}

The iCD method exhibits scale-selection sensitivity across both homogeneous (VGG13 $\rightarrow$ VGG8) and heterogeneous (ResNet32$\times$4 $\rightarrow$ ShuffleNetV1) teacher-student pairs, with optimal configurations largely invariant to architecture.
On heterogeneous architectures, multi-scale combinations (e.g., $M=\{1,2\}$ or $M=\{1,2,4\}$) yield sustained performance gains (improving KD from 61.68\% to 65.88\%), demonstrating that multi-granularity local structural knowledge enables the student to better mimic the teacher’s logits distribution.
While homogeneous architectures achieve competitive performance with a single scale ($M=\{4\}$), multi-scale fusion, especially $M=\{2,4\}$ and $M=\{1,2,4\}$, further elevates accuracy to 68.38\% and 68.52\%, respectively. This improvement validates that fine-grained multi-scale structural knowledge is essential for bridging representational gaps induced by architectural mismatches.
Crucially, in homogeneous architectures, $M=\{1,2\}$ underperforms combinations that include scale 4 (e.g., $M=\{4\}$, $M=\{2,4\}$), demonstrating that scale effectiveness is architecture-dependent: homogeneous models benefit more from fine-grained scales, while heterogeneous pairs favor medium granularity. Nevertheless, integrating the full spectrum of local structural knowledge consistently maximizes performance across architectures, confirming the necessity of fully mining the structural knowledge embedded in logits.

\textbf{Effects of gamma.}
As shown in Table \ref{tab:gamma}, \( \gamma \) controls the weight of structural knowledge distillation in the global training objective as the coefficient of the iCD loss. Experiments demonstrate that \( \gamma = 2 \) achieves optimal Top-1 accuracy, indicating that moderately strengthened structural guidance yields the best performance while both overly large and overly small values of $\gamma$ degrade it.
\begin{table}[ht]
    \centering
    \begin{tabular}{c | c c c c}
        \( \gamma \) & 1 & 2 & 3 & 4 \\
        \hline
        Acc & 65.71 & \textbf{65.88} & 65.81 & 65.36 \\
        \hline
        \( \gamma \) & 5 & 6 & 7 & 8 \\
        \hline
        Acc & 65.78 & 65.02 & 65.29 & 65.41 \\
    \end{tabular}
\caption{Performance of the iCD method with varying \( \gamma \) values on ResNet32×4 and ShuffleNetV1.}
    \label{tab:gamma}
\end{table}

\textbf{Training Efficiency.}
We evaluate the training time of state-of-the-art knowledge distillation methods to assess the efficiency of iCD. As shown in Table \ref{tab:te}, although iCD introduces additional computational steps, its impact on the training time is negligible. The training time of iCD increases by only 3–4 ms compared to SDD. This demonstrates that iCD efficiently captures structural information from the teacher model after partial decoupling.

\begin{table}[ht]
    \centering
    \begin{tabular}{c c c c}
        \hline
        Methods & KD & SD-KD & iCD-KD \\
        Times(ms) & 52 & 135 & 138 \\
        \hline
        Methods & DKD & SD-DKD & iCD-DKD \\
        Times(ms) & 53 & 133 & 136 \\
        \hline
        Methods & NKD & SD-NKD & iCD-NKD \\
        Times(ms) & 132 & 134 & 138 \\
        \hline
    \end{tabular}
    \caption{Per-batch training time on CUB-200, with ResNet32$\times$4 as teacher and MobileNetV2 as student.}
    \label{tab:te}
\end{table}

\begin{figure}[htbp]
\centering
\includegraphics[width=0.4\textwidth]{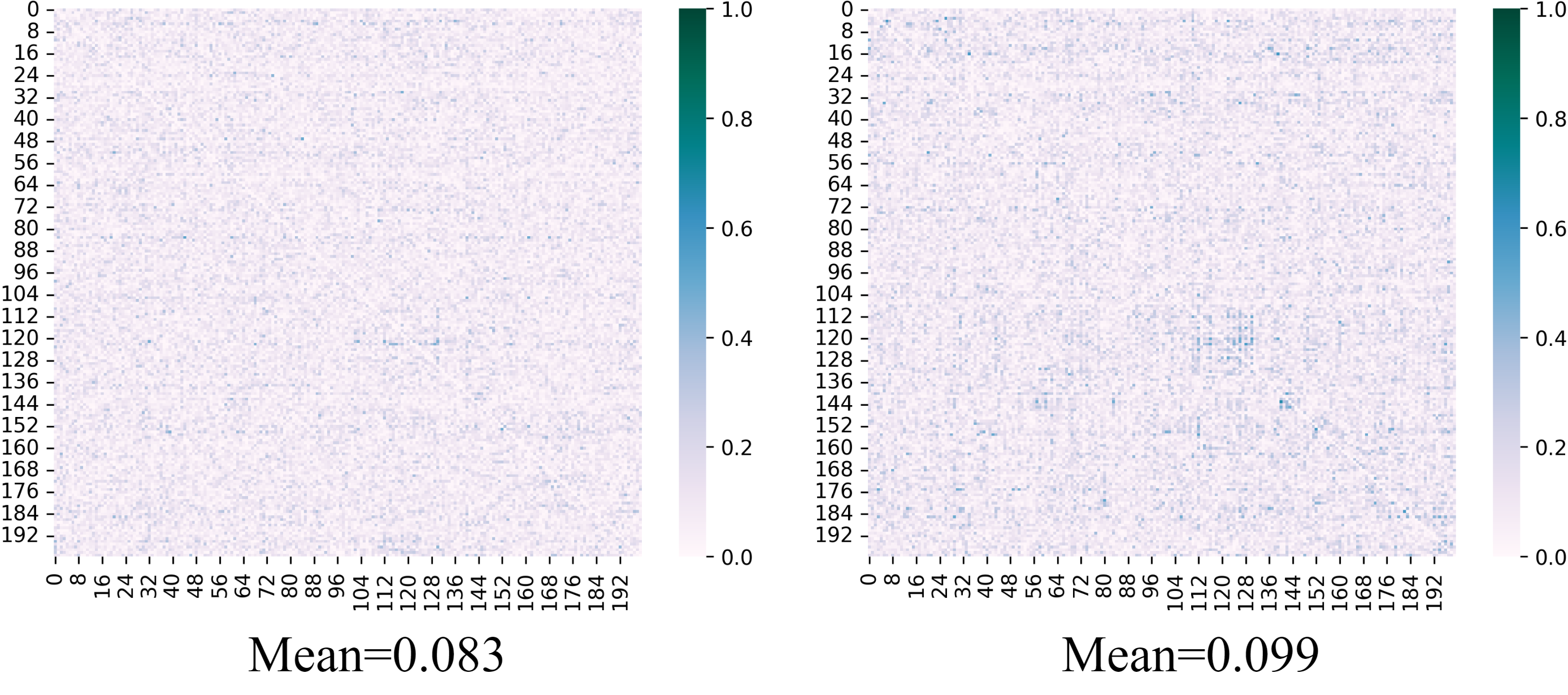} 
\caption{Discrepancy in student-teacher logits correlation matrices between KD (left) and iCD-KD (right), where "Mean" indicates the average discrepancy.}
\label{fig:map}
\end{figure}

\begin{figure}[htbp]
\centering
\includegraphics[width=0.4\textwidth]{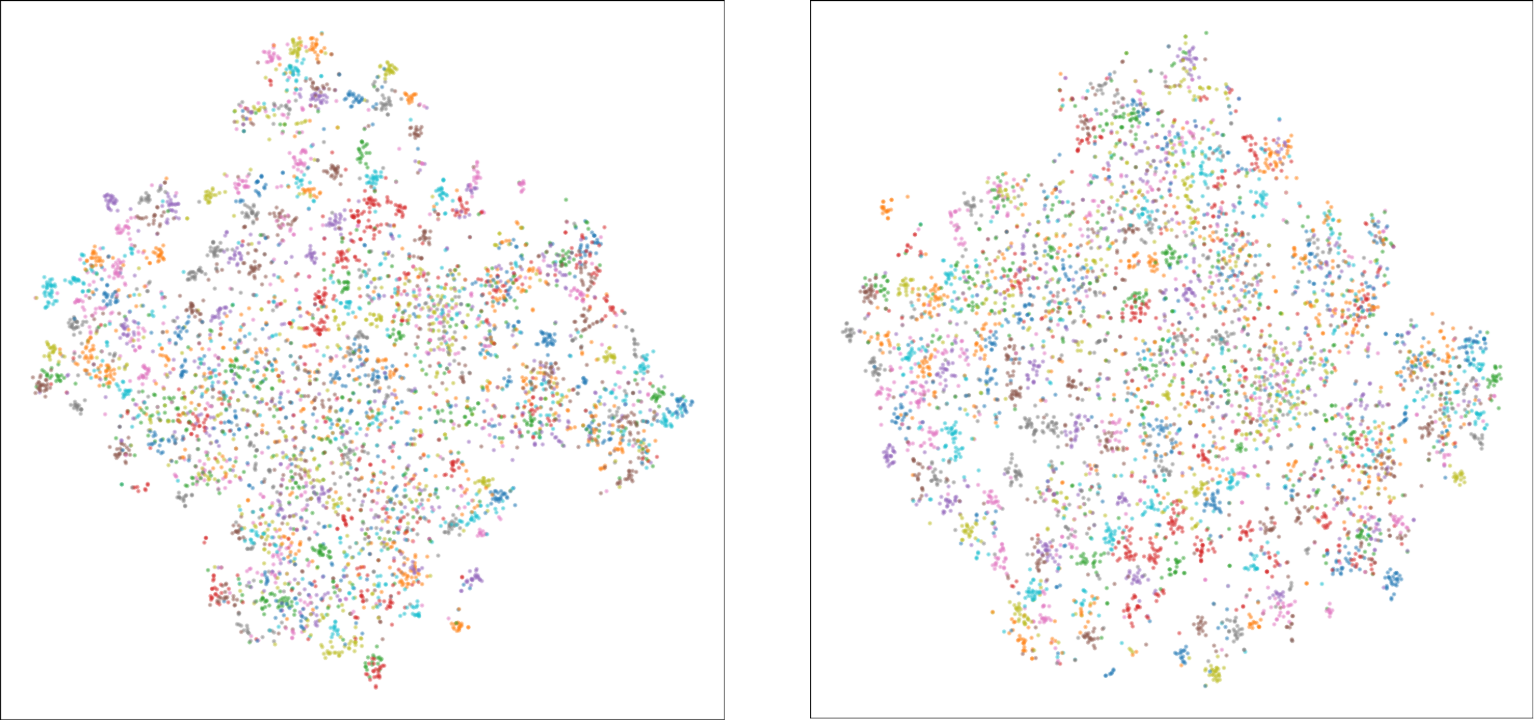} 
\caption{t-SNE visualization of features learned with KD (left) and iCD-KD (right).}
\label{fig:t-sne}
\end{figure}

\textbf{Visualization.} We conduct a visual analysis from two perspectives on the CUB-200 dataset, using ResNet32x4 as the teacher and ShuffleNetV1 as the student.
(1) Figure \ref{fig:map} visualizes the discrepancy in the global logit correlation matrices between student and teacher models. Although iCD-KD exhibits slightly larger average discrepancies than KD, it achieves higher performance. This indicates that iCD's performance gains stem not from mimicking the teacher model’s global logits but from distilled latent structural knowledge.
(2) To analyze iCD’s mechanism, Figure~\ref{fig:t-sne} visualizes feature distributions via t-SNE projections, revealing inter-class relationships. Features learned with KD exhibit significant overlap, indicating weak inter-class discrimination. In contrast, iCD-KD produces well-separated clusters with clearly defined boundaries.

\section{Conclusion}
This paper proposes iCD, a novel knowledge distillation method designed to provide student models with an interpretable form of structured guidance. While SDD delivers decoupled logit supervision, iCD further mines the latent semantic structures within teacher and student logits. By encouraging student models to implicitly mimic the teacher's semantic structures (without relying on explicit class labels), iCD achieves more structure-aware knowledge transfer. Extensive experiments demonstrate that iCD is effective across diverse teacher-student model pairs, achieving particularly strong performance in fine-grained classification tasks.
\bibliography{aaai2026}

\begin{thebibliography}{49}
\providecommand{\natexlab}[1]{#1}

\bibitem[{Ahn et~al.(2019)Ahn, Hu, Damianou, Lawrence, and
  Dai}]{ahn2019variational}
Ahn, S.; Hu, S.~X.; Damianou, A.; Lawrence, N.~D.; and Dai, Z. 2019.
\newblock Variational Information Distillation for Knowledge Transfer.
\newblock In \emph{Proceedings of the IEEE/CVF Conference on Computer Vision
  and Pattern Recognition (CVPR)}, 9155--9163.

\bibitem[{Bao et~al.(2024)Bao, Chen, Wang, Zheng, Huang, and
  Chen}]{bao2024post}
Bao, Z.; Chen, Z.; Wang, C.-D.; Zheng, W.-S.; Huang, Z.; and Chen, Y. 2024.
\newblock Post-Distillation via Neural Resuscitation.
\newblock \emph{IEEE Transactions on Multimedia}, 26: 3046--3060.

\bibitem[{Chen et~al.(2021)Chen, Liu, Zhao, and Jia}]{chen2021distilling}
Chen, P.; Liu, S.; Zhao, H.; and Jia, J. 2021.
\newblock Distilling Knowledge via Knowledge Review.
\newblock In \emph{Proceedings of the IEEE/CVF Conference on Computer Vision
  and Pattern Recognition (CVPR)}, 5006--5015.

\bibitem[{Dong, Li, and Wei(2023)}]{dong2023diswot}
Dong, P.; Li, L.; and Wei, Z. 2023.
\newblock {DisWOT}: Student Architecture Search for Distillation WithOut
  Training.
\newblock In \emph{Proceedings of the IEEE/CVF Conference on Computer Vision
  and Pattern Recognition (CVPR)}, 11898--11908.

\bibitem[{Gao et~al.(2025)Gao, Han, Zhang, Xu, Zhou, Mao, Dou, and
  Wang}]{gao2025fairness}
Gao, Z.; Han, S.; Zhang, X.; Xu, K.; Zhou, D.; Mao, X.; Dou, Y.; and Wang, H.
  2025.
\newblock Maintaining Fairness in Logit-based Knowledge Distillation for
  Class-Incremental Learning.
\newblock In \emph{Proceedings of the AAAI Conference on Artificial
  Intelligence}, volume~39, 16763--16771.

\bibitem[{Gou et~al.(2021)Gou, Yu, Maybank, and Tao}]{gou2021knowledge}
Gou, J.; Yu, B.; Maybank, S.~J.; and Tao, D. 2021.
\newblock Knowledge distillation: A survey.
\newblock \emph{International Journal of Computer Vision}, 129(6): 1789--1819.

\bibitem[{Guo et~al.(2024)Guo, Zhang, Han, Liu, Cheng, and Han}]{guo2024pixel}
Guo, G.; Zhang, D.; Han, L.; Liu, N.; Cheng, M.-M.; and Han, J. 2024.
\newblock Pixel Distillation: Cost-Flexible Distillation Across Image Sizes and
  Heterogeneous Networks.
\newblock \emph{IEEE Transactions on Pattern Analysis and Machine
  Intelligence}, 46(12): 9536--9550.

\bibitem[{Guo et~al.(2023)Guo, Yan, Li, and Lin}]{guo2023class}
Guo, Z.; Yan, H.; Li, H.; and Lin, X. 2023.
\newblock Class Attention Transfer Based Knowledge Distillation.
\newblock In \emph{Proceedings of the IEEE/CVF Conference on Computer Vision
  and Pattern Recognition (CVPR)}, 11868--11877.

\bibitem[{He et~al.(2015)He, Zhang, Ren, and Sun}]{he2015spatial}
He, K.; Zhang, X.; Ren, S.; and Sun, J. 2015.
\newblock Spatial pyramid pooling in deep convolutional networks for visual
  recognition.
\newblock \emph{IEEE Transactions on Pattern Analysis and Machine
  Intelligence}, 37(9): 1904--1916.

\bibitem[{Hinton, Vinyals, and Dean(2015)}]{hinton2015distilling}
Hinton, G.; Vinyals, O.; and Dean, J. 2015.
\newblock Distilling the Knowledge in a Neural Network.
\newblock \emph{arXiv preprint arXiv:1503.02531}.

\bibitem[{Hossain et~al.(2025)Hossain, Akhter, Hong, and
  Huh}]{hossain2025single}
Hossain, M.~I.; Akhter, S.; Hong, C.~S.; and Huh, E.-N. 2025.
\newblock Single Teacher, Multiple Perspectives: Teacher Knowledge Augmentation
  for Enhanced Knowledge Distillation.
\newblock In \emph{The Thirteenth International Conference on Learning
  Representations (ICLR)}.

\bibitem[{Huang et~al.(2023)Huang, Zhang, Zheng, You, Wang, Qian, and
  Xu}]{huang2023knowledge}
Huang, T.; Zhang, Y.; Zheng, M.; You, S.; Wang, F.; Qian, C.; and Xu, C. 2023.
\newblock Knowledge Diffusion for Distillation.
\newblock In \emph{Advances in Neural Information Processing Systems
  (NeurIPS)}, volume~36, 65299--65316.

\bibitem[{Jin, Wang, and Lin(2023)}]{jin2023multi}
Jin, Y.; Wang, J.; and Lin, D. 2023.
\newblock Multi-Level Logit Distillation.
\newblock In \emph{Proceedings of the IEEE/CVF Conference on Computer Vision
  and Pattern Recognition (CVPR)}, 24276--24285.

\bibitem[{Komodakis and Zagoruyko(2017)}]{komodakis2017paying}
Komodakis, N.; and Zagoruyko, S. 2017.
\newblock Paying more attention to attention: improving the performance of
  convolutional neural networks via attention transfer.
\newblock In \emph{International Conference on Learning Representations
  (ICLR)}.

\bibitem[{Krizhevsky, Hinton et~al.(2009)}]{krizhevsky2009learning}
Krizhevsky, A.; Hinton, G.; et~al. 2009.
\newblock Learning multiple layers of features from tiny images.

\bibitem[{Lan, Zhu, and Gong(2018)}]{lan2018knowledge}
Lan, X.; Zhu, X.; and Gong, S. 2018.
\newblock Knowledge Distillation by On-the-Fly Native Ensemble.
\newblock In \emph{Advances in Neural Information Processing Systems
  (NeurIPS)}, volume~31, 7528--7538.

\bibitem[{Li et~al.(2022)Li, Fan, Song, Li, Yunfeng, and
  Zhan}]{li2022asymmetric}
Li, X.-C.; Fan, W.-S.; Song, S.; Li, Y.; Yunfeng, S.; and Zhan, D.-C. 2022.
\newblock Asymmetric Temperature Scaling Makes Larger Networks Teach Well
  Again.
\newblock In \emph{Advances in Neural Information Processing Systems
  (NeurIPS)}, volume~35, 3830--3842.

\bibitem[{Li et~al.(2023)Li, Li, Yang, Zhao, Song, Luo, Li, and
  Yang}]{li2023curriculum}
Li, Z.; Li, X.; Yang, L.; Zhao, B.; Song, R.; Luo, L.; Li, J.; and Yang, J.
  2023.
\newblock Curriculum Temperature for Knowledge Distillation.
\newblock In \emph{Proceedings of the AAAI Conference on Artificial
  Intelligence (AAAI)}, 1504--1512.

\bibitem[{Liu et~al.(2023{\natexlab{a}})Liu, Kan, Shan, and
  Chen}]{liu2023function}
Liu, D.; Kan, M.; Shan, S.; and Chen, X. 2023{\natexlab{a}}.
\newblock Function-consistent feature distillation.
\newblock In \emph{International Conference on Learning Representations
  (ICLR)}.

\bibitem[{Liu et~al.(2023{\natexlab{b}})Liu, Li, Li, and Yao}]{liu2023norm}
Liu, X.; Li, L.; Li, C.; and Yao, A. 2023{\natexlab{b}}.
\newblock {NORM}: Knowledge Distillation via {N}-to-One Representation
  Matching.
\newblock In \emph{International Conference on Learning Representations
  (ICLR)}.

\bibitem[{Mirzadeh et~al.(2020)Mirzadeh, Farajtabar, Li, Levine, Matsukawa, and
  Ghasemzadeh}]{mirzadeh2020improved}
Mirzadeh, S.~I.; Farajtabar, M.; Li, A.; Levine, N.; Matsukawa, A.; and
  Ghasemzadeh, H. 2020.
\newblock Improved Knowledge Distillation via Teacher Assistant.
\newblock In \emph{Proceedings of the AAAI Conference on Artificial
  Intelligence (AAAI)}, 5191--5198.

\bibitem[{M{\"u}ller, Kornblith, and Hinton(2019)}]{muller2019when}
M{\"u}ller, R.; Kornblith, S.; and Hinton, G.~E. 2019.
\newblock When Does Label Smoothing Help?
\newblock In \emph{Advances in Neural Information Processing Systems
  (NeurIPS)}, volume~32.

\bibitem[{Park et~al.(2019)Park, Kim, Lu, and Cho}]{park2019relational}
Park, W.; Kim, D.; Lu, Y.; and Cho, M. 2019.
\newblock Relational Knowledge Distillation.
\newblock In \emph{Proceedings of the IEEE/CVF Conference on Computer Vision
  and Pattern Recognition (CVPR)}, 3962--3971.

\bibitem[{Romero et~al.(2015)Romero, Ballas, Kahou, Chassang, Gatta, and
  Bengio}]{romero2015fitnets}
Romero, A.; Ballas, N.; Kahou, S.~E.; Chassang, A.; Gatta, C.; and Bengio, Y.
  2015.
\newblock FitNets: Hints for thin deep nets.
\newblock In \emph{International Conference on Learning Representations
  (ICLR)}.

\bibitem[{Shen et~al.(2021)Shen, Liu, Xu, Chen, Cheng, and
  Savvides}]{shen2021label}
Shen, Z.; Liu, Z.; Xu, D.; Chen, Z.; Cheng, K.~T.; and Savvides, M. 2021.
\newblock Is Label Smoothing Truly Incompatible with Knowledge Distillation: An
  Empirical Study.
\newblock In \emph{International Conference on Learning Representations
  (ICLR)}.

\bibitem[{Singh and Wang(2024)}]{singh2024simple}
Singh, A.; and Wang, H. 2024.
\newblock Simple unsupervised knowledge distillation with space similarity.
\newblock In \emph{Proceedings of the European Conference on Computer Vision
  (ECCV)}, 147--164.

\bibitem[{Son et~al.(2021)Son, Na, Choi, and Hwang}]{son2021densely}
Son, W.; Na, J.; Choi, J.; and Hwang, W. 2021.
\newblock Densely Guided Knowledge Distillation Using Multiple Teacher
  Assistants.
\newblock In \emph{Proceedings of the IEEE/CVF International Conference on
  Computer Vision (ICCV)}, 9395--9404.

\bibitem[{Sun et~al.(2024)Sun, Ren, Li, Wang, and Cao}]{sun2024logit}
Sun, S.; Ren, W.; Li, J.; Wang, R.; and Cao, X. 2024.
\newblock Logit Standardization in Knowledge Distillation.
\newblock In \emph{Proceedings of the IEEE/CVF Conference on Computer Vision
  and Pattern Recognition (CVPR)}, 15731--15740.

\bibitem[{Tang et~al.(2020)Tang, Shivanna, Zhao, Lin, Singh, Chi, and
  Jain}]{tang2020understanding}
Tang, J.; Shivanna, R.; Zhao, Z.; Lin, D.; Singh, A.; Chi, E.~H.; and Jain, S.
  2020.
\newblock Understanding and improving knowledge distillation.
\newblock \emph{arXiv preprint arXiv:2002.03532}.

\bibitem[{Tian, Krishnan, and Isola(2020)}]{tian2020contrastive}
Tian, Y.; Krishnan, D.; and Isola, P. 2020.
\newblock Contrastive representation distillation.
\newblock In \emph{International Conference on Learning Representations
  (ICLR)}.

\bibitem[{Tung and Mori(2019)}]{tung2019similarity}
Tung, F.; and Mori, G. 2019.
\newblock Similarity-Preserving Knowledge Distillation.
\newblock In \emph{Proceedings of the IEEE/CVF International Conference on
  Computer Vision (ICCV)}, 1365--1374.

\bibitem[{Wang et~al.(2023)Wang, Chen, Mei, Zhang, Feng, and
  Chen}]{wang2023semckd}
Wang, C.; Chen, D.; Mei, J.-P.; Zhang, Y.; Feng, Y.; and Chen, C. 2023.
\newblock {SemCKD}: Semantic Calibration for Cross-Layer Knowledge
  Distillation.
\newblock \emph{IEEE Transactions on Knowledge and Data Engineering}, 35(6):
  6305--6319.

\bibitem[{Wei, Luo, and Luo(2024)}]{wei2024scale}
Wei, S.; Luo, C.; and Luo, Y. 2024.
\newblock Scale Decoupled Distillation.
\newblock In \emph{Proceedings of the IEEE/CVF Conference on Computer Vision
  and Pattern Recognition (CVPR)}, 15975--15983.

\bibitem[{Wei et~al.(2024)Wei, Luo, Luo, and Xu}]{wei2024privileged}
Wei, S.; Luo, C.; Luo, Y.; and Xu, J. 2024.
\newblock Privileged Modality Learning via Multimodal Hallucination.
\newblock \emph{IEEE Transactions on Multimedia}, 26: 1516--1527.

\bibitem[{Welinder et~al.(2010)Welinder, Branson, Mita, Wah, Schroff, Belongie,
  and Perona}]{Welinder2010}
Welinder, P.; Branson, S.; Mita, T.; Wah, C.; Schroff, F.; Belongie, S.; and
  Perona, P. 2010.
\newblock Caltech-ucsd birds 200.

\bibitem[{Xu et~al.(2020{\natexlab{a}})Xu, Liu, Li, and Loy}]{xu2020knowledge}
Xu, G.; Liu, Z.; Li, X.; and Loy, C.~C. 2020{\natexlab{a}}.
\newblock Knowledge Distillation Meets Self-Supervision.
\newblock In \emph{Proceedings of the European Conference on Computer Vision
  (ECCV)}, 588--604.

\bibitem[{Xu et~al.(2020{\natexlab{b}})Xu, Rui, Li, and Gu}]{xu2020feature}
Xu, K.; Rui, L.; Li, Y.; and Gu, L. 2020{\natexlab{b}}.
\newblock Feature Normalized Knowledge Distillation for Image Classification.
\newblock In \emph{Proceedings of the European Conference on Computer Vision
  (ECCV)}, 664--680.

\bibitem[{Xu et~al.(2025)Xu, Liu, Liu, Wang, Xu, and Cheng}]{xu2025local}
Xu, L.; Liu, K.; Liu, J.; Wang, L.; Xu, L.; and Cheng, J. 2025.
\newblock Local Dense Logit Relations for Enhanced Knowledge Distillation.
\newblock \emph{arXiv preprint arXiv:2507.15911}.

\bibitem[{Xu et~al.(2024)Xu, Ren, Huang, Zheng, and Chen}]{xu2024improving}
Xu, L.; Ren, J.; Huang, Z.; Zheng, W.; and Chen, Y. 2024.
\newblock Improving Knowledge Distillation via Head and Tail Categories.
\newblock \emph{IEEE Transactions on Circuits and Systems for Video
  Technology}, 34(5): 3465--3480.

\bibitem[{Xue et~al.(2021{\natexlab{a}})Xue, Song, Wang, Chen, Wang, and
  Song}]{ijcai2021p444}
Xue, M.; Song, J.; Wang, X.; Chen, Y.; Wang, X.; and Song, M.
  2021{\natexlab{a}}.
\newblock KDExplainer: A Task-oriented Attention Model for Explaining Knowledge
  Distillation.
\newblock In \emph{Proceedings of the International Joint Conference on
  Artificial Intelligence (IJCAI)}, 3228--3234.

\bibitem[{Xue et~al.(2021{\natexlab{b}})Xue, Song, Wang, Chen, Wang, and
  Song}]{xue2021kdexplainer}
Xue, M.; Song, J.; Wang, X.; Chen, Y.; Wang, X.; and Song, M.
  2021{\natexlab{b}}.
\newblock {KDExplainer}: A Task-oriented Attention Model for Explaining
  Knowledge Distillation.
\newblock In \emph{Proceedings of the Thirtieth International Joint Conference
  on Artificial Intelligence (IJCAI)}, 3228--3234.

\bibitem[{Yang et~al.(2024)Yang, Li, Zeng, Li, Yuan, and Li}]{yang2024vittkd}
Yang, Z.; Li, Z.; Zeng, A.; Li, Z.; Yuan, C.; and Li, Y. 2024.
\newblock {ViTKD}: Feature-Based Knowledge Distillation for Vision
  Transformers.
\newblock In \emph{Proceedings of the IEEE/CVF Conference on Computer Vision
  and Pattern Recognition Workshops (CVPR Workshops)}, 1379--1388.

\bibitem[{Yang et~al.(2023)Yang, Zeng, Li, Zhang, Yuan, and Li}]{yang2023from}
Yang, Z.; Zeng, A.; Li, Z.; Zhang, T.; Yuan, C.; and Li, Y. 2023.
\newblock From Knowledge Distillation to Self-Knowledge Distillation: A Unified
  Approach with Normalized Loss and Customized Soft Labels.
\newblock In \emph{Proceedings of the IEEE/CVF International Conference on
  Computer Vision (ICCV)}, 17139--17148.

\bibitem[{Yao and Sun(2020)}]{yao2020knowledge}
Yao, A.; and Sun, D. 2020.
\newblock Knowledge Transfer via Dense Cross-Layer Mutual-Distillation.
\newblock In \emph{Proceedings of the European Conference on Computer Vision
  (ECCV)}, 294--311.

\bibitem[{Yim et~al.(2017)Yim, Joo, Bae, and Kim}]{yim2017gift}
Yim, J.; Joo, D.; Bae, J.; and Kim, J. 2017.
\newblock A Gift from Knowledge Distillation: Fast Optimization, Network
  Minimization and Transfer Learning.
\newblock In \emph{Proceedings of the IEEE Conference on Computer Vision and
  Pattern Recognition (CVPR)}, 7130--7138.

\bibitem[{Zhang et~al.(2018)Zhang, Xiang, Hospedales, and Lu}]{zhang2018deep}
Zhang, Y.; Xiang, T.; Hospedales, T.~M.; and Lu, H. 2018.
\newblock Deep Mutual Learning.
\newblock In \emph{Proceedings of the IEEE/CVF Conference on Computer Vision
  and Pattern Recognition (CVPR)}, 4320--4328.

\bibitem[{Zhao et~al.(2022)Zhao, Cui, Song, Qiu, and Liang}]{zhao2022decoupled}
Zhao, B.; Cui, Q.; Song, R.; Qiu, Y.; and Liang, J. 2022.
\newblock Decoupled Knowledge Distillation.
\newblock In \emph{Proceedings of the IEEE/CVF Conference on Computer Vision
  and Pattern Recognition (CVPR)}, 11943--11952.

\bibitem[{Zheng and Yang(2024)}]{zheng2024knowledge}
Zheng, K.; and Yang, E.-H. 2024.
\newblock Knowledge Distillation Based on Transformed Teacher Matching.
\newblock In \emph{International Conference on Learning Representations
  (ICLR)}.

\bibitem[{Zhou et~al.(2021)Zhou, Song, Chen, Zhou, Wang, Yuan, and
  Qian}]{zhou2021rethinking}
Zhou, H.; Song, L.; Chen, J.; Zhou, Y.; Wang, G.; Yuan, J.; and Qian, Z. 2021.
\newblock Rethinking Soft Labels for Knowledge Distillation: A Bias-Variance
  Tradeoff Perspective.
\newblock In \emph{International Conference on Learning Representations
  (ICLR)}.

\end{thebibliography}

\end{document}